\documentclass[letterpaper, 10 pt, conference]{ieeeconf} 
\IEEEoverridecommandlockouts
\usepackage{mathrsfs}
\usepackage{xcolor}

\usepackage{graphics} 
\usepackage{epsfig} 
\usepackage{amsmath} 
\usepackage{amssymb} 
\usepackage{color}
\usepackage{bm}
\usepackage{subcaption}
\usepackage{multirow}
\usepackage{array}
\usepackage{booktabs}
\usepackage{algorithm,algcompatible}
\usepackage{mathtools}
\usepackage{cite}
\usepackage{hyperref}
\usepackage{cuted}
\definecolor{pointcolor}{RGB}{0,114,189}
\definecolor{linecolor}{RGB}{217,83,25}
\definecolor{planecolor}{RGB}{237,177,32}
\definecolor{spherecolor}{RGB}{119,172,48}
\definecolor{ellipsoidcolor}{RGB}{126,47,142}
\definecolor{cylindercolor}{RGB}{77,190,238}
\definecolor{conecolor}{RGB}{162,20,47}

\graphicspath{{figures/}}

\title{\huge \bf FE-DeTr: Keypoint Detection and Tracking in Low-quality Image Frames with Events }
\author{
Xiangyuan Wang$^*$ \quad Kuangyi Chen$^*$ \quad  Wen Yang \quad Lei Yu \quad Yannan Xing \quad Huai Yu$^\dagger$
\thanks{$^*$Equal contribution. $^\dagger$Corresponding author.}
\thanks{X. Wang, K. Chen,  W. Yang, L. Yu and H. Yu are with the EIS, Wuhan University, Wuhan, China {\tt\small\{wangxiangyuan,chenky721, yangwen,ly.wd,yuhuai\}@whu.edu.cn}. Y. Xing is with the SynSense Tech. Co. Ltd., Chengdu, China {\tt\small yannan.xing@synsense.ai}.}
}

\begin{document}

\maketitle
\begin{abstract}

Keypoint detection and tracking in traditional image frames are often compromised by image quality issues such as motion blur and extreme lighting conditions. Event cameras offer potential solutions to these challenges by virtue of their high temporal resolution and high dynamic range. However, they have limited performance in practical applications due to their inherent noise in event data. This paper advocates fusing the complementary information from image frames and event streams to achieve more robust keypoint detection and tracking. Specifically, we propose a novel keypoint detection network that fuses the textural and structural information from image frames with the high-temporal-resolution motion information from event streams, namely FE-DeTr. The network leverages a temporal response consistency for supervision, ensuring stable and efficient keypoint detection. Moreover, we use a spatio-temporal nearest-neighbor search strategy for robust keypoint tracking.
Extensive experiments are conducted on a new dataset featuring both image frames and event data captured under extreme conditions. The experimental results confirm the superior performance of our method over both existing frame-based and event-based methods. Our code, pre-trained models, and dataset are available at \url{https://github.com/yuyangpoi/FE-DeTr}.

\end{abstract}
\section{Introduction}

Keypoint detection and tracking serve as critical components for a range of applications, such as Simultaneous Localization and Mapping (SLAM) and Structure from Motion (SfM). 
Traditional frame-based methods \cite{detone2018superpoint, christiansen2019unsuperpoint, revaud2019r2d2, dosovitskiy2015flownet, teed2020raft, tian2017l2, jiang2021cotr, doersch2022tap, harley2022particle, doersch2023tapir} rely on sharp, distinct features within an image to detect and track keypoints. 
However, motion blur substantially distorts these features, making keypoints difficult to locate.
Additionally, motion information present during the exposure time can't be captured, making accurate keypoint tracking an insurmountable challenge. Extreme lighting conditions, such as overexposure and low light, exacerbate these challenges by producing frames that are either washed out or inadequately illuminated, further hampering feature identification. Although BALF\cite{zhao2022balf} introduces an MLP-based architecture for local keypoint detection within blurred images, it still can't effectively handle images under extreme lighting conditions.

Event cameras can overcome these limitations because they capture environmental changes asynchronously, providing events with extraordinarily high temporal resolution and high dynamic range \cite{gallego2020event, zhang2023generalizing, chen2024motion}. These characteristics enable them to capture necessary information for keypoint detection and tracking \cite{vasco2016fast, mueggler2017fast, alzugaray2018asynchronous, li2019fa
, manderscheid2019speed, chiberre2021detecting, chiberre2022long, zhu2017event, gehrig2018asynchronous, messikommer2023data} under challenging conditions.
Nonetheless, event cameras also come with their limitations: they only capture dynamic changes in luminance, lack details on absolute intensity or texture, and are noisy due to both operational principles and hardware constraints.
On the contrary, image frames preserve stable structural features, and also mitigate the impact of event noise \cite{yu2023detecting}.
Motivated by the complementary strengths of both, we propose fusing image frames with event data to enhance keypoint detection and tracking performance in challenging conditions.

\begin{figure}[t]
    \centering
    \includegraphics[width=1.0\linewidth]{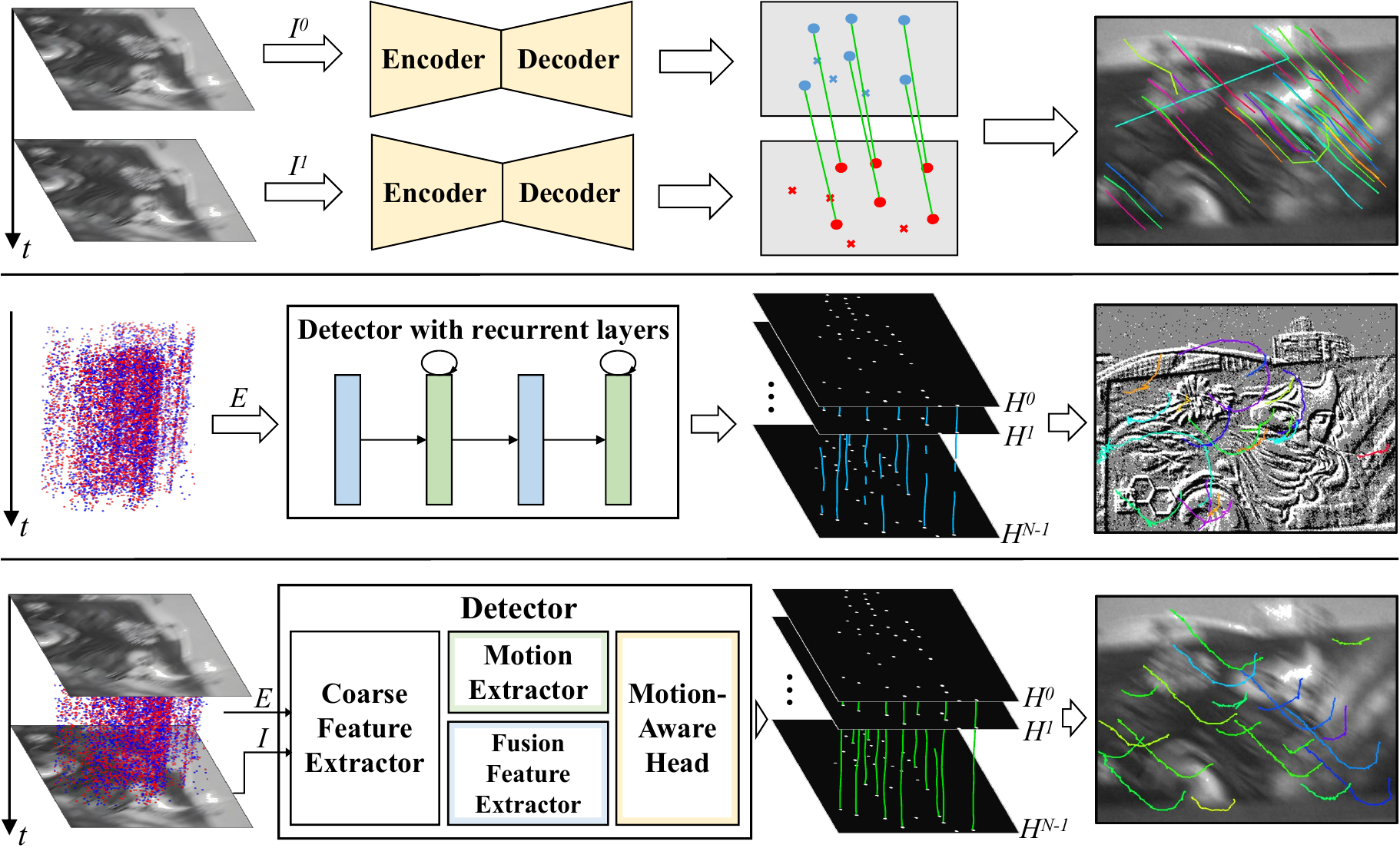}
    \caption{Our method (\textbf{bottom}) leverages the complementary characteristics of image frames and event streams, allowing for stable keypoint detection and tracking in extreme conditions compared to frame-based methods (\textbf{top}) and event-based methods (\textbf{middle}).} \vspace{-5mm}
    \label{fig:Pipeline}
\end{figure}

In this paper, we propose a keypoint detection and tracking framework that fuses image frames and event data (\textit{i.e.}, FE-DeTr).
The detection network in the framework is trained using a supervision strategy based on temporal response consistency. 
The objective is to exploit the complementary information from image frames and event data to identify keypoints exhibiting temporal and spatial persistence, enabling their extended tracking using a spatio-temporal nearest-neighbor strategy. 
An overall diagram of our proposed approach is illustrated in Fig. \ref{fig:Pipeline}.
The keypoint detection network comprises a Fusion Feature Extractor (FFE), a Motion Extractor (ME), and a Motion-Aware Head (MAH). The FFE fuses complementary information from image frames and events and achieves the propagation of temporal information through a recurrent structure. The ME includes spatial and channel attention mechanisms, designed to extract motion information from event streams. The MAH incorporates an iterative strategy for handling relative motion, implicitly warping keypoint responses at different time instants using deformable convolutions to improve response repeatability. The supervision based on temporal response consistency relies solely on the relative motion relationships between different time instants, further ensuring the repeatability of detected keypoints.


The main contributions are listed as follows:

\begin{itemize}
    \item We propose the first framework fusing image frames and events for robust keypoint detection and tracking under extreme conditions.
    \item 
    We design a Motion-Aware Head based on an iterative strategy and introduce a supervision strategy built on temporal response consistency, which enables the network to produce stable and highly repeatable responses for long-term keypoint tracking.
    \item We contribute a new keypoint detection and tracking dataset that contains both image frames and event data, encompassing high-speed motion and extreme lighting scenarios. Experimental results on this dataset demonstrate our method outperforms both frame-based and event-based methods under extreme conditions.
\end{itemize}

\section{Related Work}
\label{sec:related_work}
In this section, we review the frame-based and event-based methods for keypoint detection and tracking.  

\vspace{-0.3mm}
\subsection{Frame-Based methods} 
Learning-based keypoint detection and tracking on image frames have been studied over years. Most keypoint detection methods also encode local descriptors for matching.

\textbf{Keypoint detection}:
Lift \cite{yi2016lift} is an early pioneer in exploring CNNs for local feature extraction and description.
Superpoint \cite{detone2018superpoint} presents a learning-based local descriptor and a self-supervised keypoint detection that utilizes pseudo-ground truth correspondences with homographic transformations.
Unsuperpoint \cite{christiansen2019unsuperpoint} and R2D2 \cite{revaud2019r2d2} propose to extract highly repeatable keypoints from image geometric transformations.
SiLK \cite{gleize2023silk} summarizes prior work and improves the stability of detection and matching by defining transition probabilities on descriptor similarity.
BALF\cite{zhao2022balf} utilizes an MLP-based structure to effectively detect keypoint within blurred images. However, on the one hand, the definition of keypoints in blurred images isn't well clarified. On the other hand, it is only effective for handling blurred images under favorable lighting conditions.
The aforementioned methods inspire the design of our approach, yet none of them explicitly tackles the challenges posed by low-quality images.

\textbf{Point tracking}:
Point tracking aims to establish point-level correspondences across multiple images. Currently, mainstream methods encompass optical flow, descriptor-based matching, and particle video techniques. Optical flow methods \cite{lucas1981iterative,dosovitskiy2015flownet,sun2018pwc,teed2020raft} estimate point correspondences between continuous frames based on the intensity invariance over time, limiting their robustness for long-term tracking. Descriptor-based matching techniques \cite{yi2016lift,detone2018superpoint,christiansen2019unsuperpoint,revaud2019r2d2,gleize2023silk} overlook motion information between consecutive frames, instead identifying point correspondences based on descriptors across image pairs. While particle video methods offer high accuracy \cite{doersch2022tap,harley2022particle,doersch2023tapir}, they require multiple input frames and consequently lack real-time performance capabilities. Similarly, all these point-tracking methods depend on high-quality images and are thus sensitive to motion blur and extreme lighting conditions.

\subsection{Event-Based methods} 
In recent years, there has been a growing trend to leverage event cameras for enhanced keypoint detection and tracking.
However, varied motions elicit unique event responses, posing challenges for early handcrafted methods \cite{vasco2016fast, mueggler2017fast, alzugaray2018asynchronous, li2019fa} in robustly extracting keypoints. As a result, learning-based techniques have gained prominence. SILC \cite{manderscheid2019speed} employs random forests for keypoint localization and introduces Speed-Invariant Time-Surface to mitigate the influence of motion speed on event representations.
Inspired by the image gradient-based corner detectors, Gradient \cite{chiberre2021detecting} proposes to reconstruct gradient maps from events for keypoint detection.
Similarly, Long-lived\cite{chiberre2022long} adopts the same network architecture as \cite{chiberre2021detecting} but directly predicts keypoint heatmaps instead of image gradients, achieving superior performance.
All the above methods achieve high-frequency keypoint detection and tracking.
However, relying solely on event-based methods proves challenging for mitigating inherent sensor noise, thereby constraining the effectiveness of keypoint detection and tracking in event streams. In this paper, we fuse traditional cameras with event cameras to combine the strengths of each modality, further enhancing the robustness of keypoint detection and tracking.

\section{Method}
\label{sec:Method}


\subsection{Event representaion}
\begin{figure*}[t]
    \centering
    \includegraphics[width=0.95\linewidth]{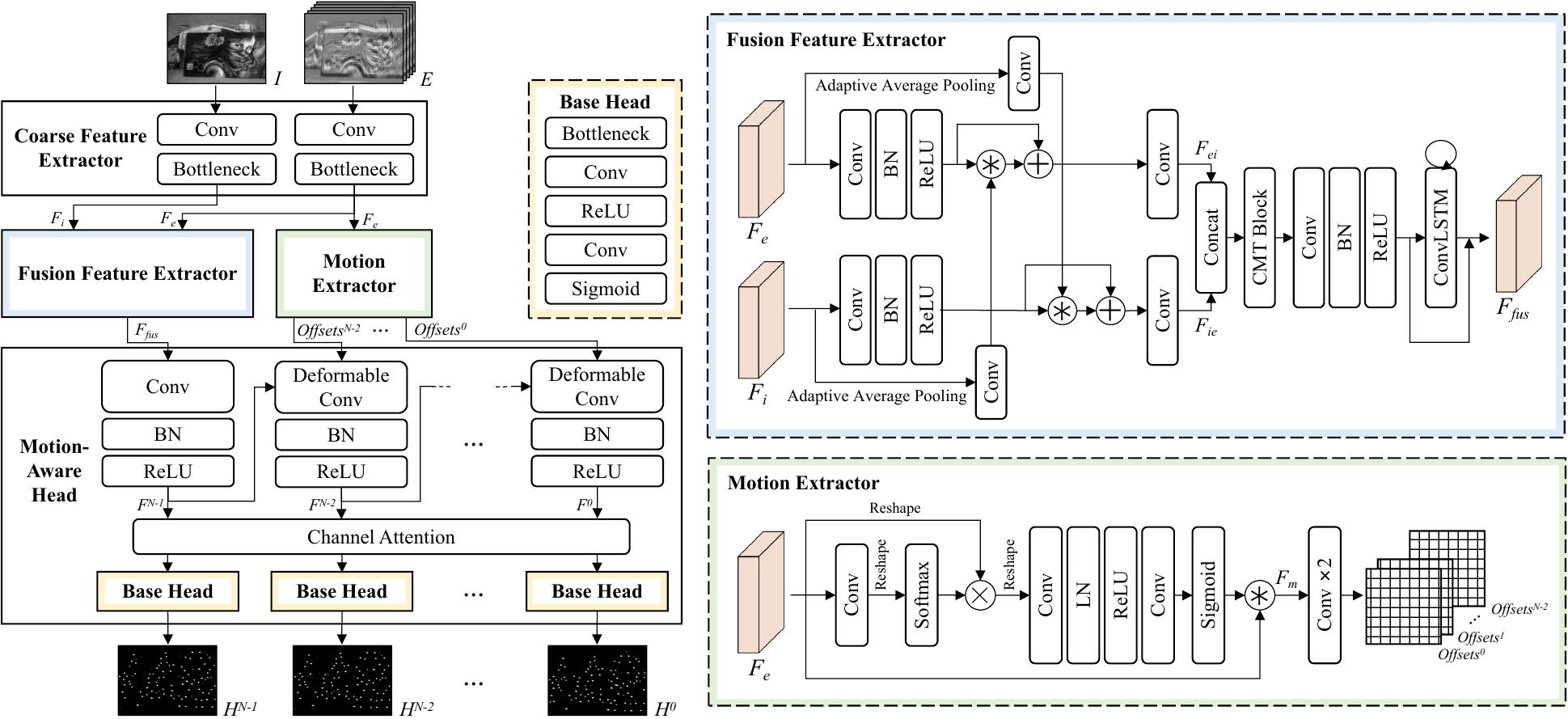}
    \caption{Overview of the proposed FE-DeTr. For each frame interval, an event representation is generated and combined with the image frame as input to the keypoint detection network. The network outputs a sequence of uniformly spaced heatmaps.} \vspace{-5mm}
    \label{fig:Network_Architecture}
\end{figure*}

To input asynchronous events to the neural network, we employ the Voxel Grid \cite{zhu2019unsupervised} as the event representation $E$. 
This event representation effectively preserves the temporal information of events. 
For a event $(x_i, y_i, t_i, p_i)$ within a specific frame interval, the polarity $p_i$ is allocated to the two temporally nearest bins in $E$ as following:
\begin{equation}
\begin{split}
E(x, y, t) &= \sum_{i}p_i\max(0, 1-\vert t-t_i^* \vert),\\
t_i^* &= \frac{(t_i - t_{min})}{t_{max} - t_{min}}(B-1),
\end{split}
\end{equation}
where $x$, $y$ and $t$ are the coordinates in the \textit{x-y-time} dimensions. $t_{min}$ and $t_{max}$ are the beginning and end times of the frame interval, which is temporally divided into $B$ bins.

\subsection{Network Architecture}
As shown in Fig. \ref{fig:Network_Architecture}, we first extract coarse features from image frame $I$ and event Voxel Grid $E$, respectively. Subsequently, the image feature $F_i$ and event feature $F_e$ are fed into the Fusion Feature Extractor (FFE). Within the FFE, a unified feature $F_{fus}$ is generated, with a focus on emphasizing features at the end of the exposure time. This fused feature effectively combines information from two modalities.
The event features $F_e$ are fed into the Motion Extractor (ME) to extract motion information, represented as a set of offsets $\mathcal{O}$ = \{$\mathit{Offset}^{N-2}$, $\mathit{Offset}^{N-3}$, ..., $\mathit{Offset}^0$\}.
This motion information is subsequently supplied to the Motion-Aware Head (MAH), which performs implicit warping of $F_{fus}$ to different time instants within the frame interval and produces a set of heatmaps \{$H^0$, $H^1$, ..., $H^{N-1}$\} corresponding to each instant. 
Here, $N$ represents the number of output heatmaps, with a larger $N$ resulting in higher temporal resolution for detection. The entire process is designed to produce more stable detection results, which is also the key to achieving long-term tracking. 

\subsubsection{\textbf{FFE}}
To adaptively extract complementary information from images and events, we employ dynamic filters to enhance one modality with respect to the other.
The process is formulated as:
\begin{equation}
\begin{split}
    F_{ei} &= \mathrm{Conv_{1 \times 1}}(\mathcal{K}_i \circledast \mathcal{F}   + \mathcal{F}),\\
    \mathcal{K}_i &= \mathrm{Conv_{3 \times 3}}(\mathcal{A}(F_i)),\\
    \mathcal{F} &= \mathrm{ReLU}(\mathrm{BN}(\mathrm{Conv_{3 \times 3}}(F_e))),\\
\end{split}
\label{dynamic filter}
\end{equation}
where $\circledast$ represents depthwise convolution, and $\mathcal{A}$ signifies adaptive average pooling. 
It is worth noting that Eq. (\ref{dynamic filter}) is symmetric when processing both the event and image frame.

After concatenating $F_{ei}$ and $F_{ie}$, the combined features are passed through the CMT Block \cite{guo2022cmt}, which can realize global inter-modal feature interactions and information fusion. Subsequently, a combination of convolution, Batch Normalization (BN), and Rectified Linear Unit (ReLU) is applied for the adjustment of the channel dimensions.
The features are then processed by a ConvLSTM layer \cite{shi2015convolutional}, enhanced with residual connections, to produce the final fused feature $F_{fus}$.
Meanwhile, ConvLSTM handles the propagation of temporal information, ensuring that the network produces temporally consistent results. 

\subsubsection{\textbf{ME}}
The Motion Extractor is responsible for extracting motion information from the high-temporal-resolution events. We utilized the MA module \cite{zhang2023frame} to extract motion features, denoted as $F_m$. The MA module employs spatial and channel attention mechanisms to enhance valuable information within the event modality, simultaneously emphasizing motion cues while suppressing noise. Subsequently, $F_m$ is passed through a two-layer convolution to obtain the required offsets for deformable convolution in MAH. 

\subsubsection{\textbf{MAH}}
The Motion-Aware Head is responsible for propagating the stable fusion feature $F_{fus}$ to various time instants within the frame interval using motion information $\mathcal{O}$.
The module employs an iterative strategy that incorporates deformable convolution, which can achieve implicit warping.

We first extract the feature $F^{N-1}$ aligned with the end of the exposure time.
Subsequently, we employ iterative steps through a combination of deformable convolution, BN, and ReLU to propagate $F^{N-1}$ to other time instants, therefore obtain features \{$F^{N-2}$, $F^{N-3}$, ..., $F^0$\} for different time instants. This process can be formulated as follows:
\begin{equation}
\begin{split}
    F^{N-1} &= \mathrm{ReLU}(\mathrm{BN}(\mathrm{Conv_{3 \times 3}}(F_{fus}))),\\
    F^{n-1} &= \mathrm{ReLU}(\mathrm{BN}(\mathrm{DConv_{3 \times 3}}(F^{n}, \mathit{Offset}^{n-1}))),\\
\end{split}
\label{MAH output}
\end{equation}
where $n \in \mathbb{Z},\, 1 \leq n \leq N-1$.

After applying channel attention \cite{hu2018squeeze}, the features \{$F^{N-1}$, $F^{N-2}$, ..., $F^0$\} are fed into the Base Head, which is composed of several convolutional structures with a Sigmoid function to normalize the heatmap outputs. As a result, a set of keypoint heatmaps \{$H^{N-1}$, $H^{N-2}$, ..., $H^0$\} are generated. 

\subsection{Loss Function}
Generally, pseudo-labels of keypoints are generated on image frames with Harris or SIFT features \cite{chiberre2022long}.
These pseudo-labels are subsequently used to supervise event-based methods. Although this kind of supervision is straightforward, it essentially mimics handcrafted detectors without considering their applicability to event data. Additionally, these methods care little about the keypoint's repeatability.

Inspired by the self-supervised frame-based keypoint detection methods \cite{detone2018superpoint, christiansen2019unsuperpoint, revaud2019r2d2}, we employ a loss function $L_{consist}$ based on temporal response consistency to supervise the network. It relies solely on the image transformation between different time instants.
We achieve this by utilizing temporally-sequenced heatmaps generated by our network and comparing them at various time instances.
This loss function encourages the network to identify keypoint locations that remain stable across the entire temporal axis, while also minimizing the variance of responses at these locations.
It is expressed as follows: 
\begin{equation}
    L_{consist} = 1 - \frac{1}{\vert \mathcal{P} \vert}\sum_{p \in \mathcal{P}} \mathcal{C}[p],
\label{Consistency loss}
\end{equation}
\begin{equation}
\mathcal{C}[p] = \mathrm{Dist}(H^n[p], \mathrm{Warp}(H^{n-1}, \mathcal{T}^{(n-1, n)})[p]),
\label{Consist}
\end{equation}
where $\mathcal{P}=\{p\}$ represents a collection of $M \times M$ patches extracted from the image region. $\mathrm{Dist}$ is a distance metric such as cosine similarity or L1 distance that reflects the difference between images. And $\mathrm{Warp}(*, \mathcal{T}^{(a, b)})$ denotes the operation of warping the image from time $a$ to time $b$ using the image correspondence $\mathcal{T}^{(a, b)}$. $\mathcal{C}[p]$ signifies the magnitude of consistency of patch $p$.

$L_{consist}$ tends to flatten the response values between each patch, resulting in a large and smooth response on the network output. Thus, following \cite{revaud2019r2d2}\cite{zhang2018learning}, we incorporate an additional loss function $L_{peaky}$ for assistance:
\begin{equation}
\begin{split}
    L_{peaky} =  1-\frac{1}{\vert \mathcal{P} \vert}\sum_{p \in \mathcal{P}}(\max H^n[p] - \mathop{{\rm mean}}\limits H^n[p]).
\end{split}
\label{Peaky loss}
\end{equation}

$L_{peaky}$ stretches the responses between local peaks and local averages, thereby preventing local smoothing. However, $L_{peaky}$ tends to generate responses on each patch, which does not align with the reality of the situation.
Specifically, intensity-homogeneous areas are not informative for keypoint detection and tracking.
Therefore, building upon the $L_{peaky}$, we propose a Consistency Peaky Loss $L_{cp}$ that includes a consistency mask to address this issue. In intensity-homogeneous regions, the responses in heatmaps tend to be uniform, which does not align with temporal consistency. In other words, the responses transforming from time $a$ to time $b$ do not align with the responses at time $b$, resulting in a large distance metric value. Based on this observation, we suppress the responses in regions with large distance metrics while simultaneously enhancing the peaks in other regions: 
\begin{equation}
\begin{split}
    L_{cp} = 1 &- \frac{1}{\vert \mathcal{P} \vert}\sum_{p \in \mathcal{P}}(\max H^n[p] - \mathop{{\rm mean}}\limits H^n[p])\mathcal{M}[p]\\
    &+ \frac{1}{\vert \mathcal{P} \vert}\sum_{p \in \mathcal{P}}\mathop{{\rm mean}}\limits H^n[p](1-\mathcal{M}[p]),
\end{split}
\end{equation}
\begin{equation}
    \mathcal{M} = \frac{\mathcal{C} - \min(\mathcal{C})}{\max(\mathcal{C})-\min(\mathcal{C})},
\label{loss mask}
\end{equation}
where $\mathcal{C}$ is the consistency magnitude defined in Eq. (\ref{Consist}). $L_{peaky}$ can be viewed as a special case of $L_{cp}$ with $\mathcal{M}=1$.

The overall loss function is a weighted sum of the consistency loss and the consistency peak loss,
\begin{equation}
    L = L_{consist} + \alpha L_{cp}.
\label{overall loss function}
\end{equation}

The specific setting of $\alpha$ is detailed in the section \ref{sec: Training details}.

\subsection{Implementation Details}
\label{sec: Implementation Details}
\subsubsection{Spatio-Temporal Nearest-Neighbor Tracking}
Benefiting from the high temporal resolution of events, our keypoint network generates a sequence of heatmaps at evenly-spaced time intervals within the frame interval. We identify positions on these heatmaps with values exceeding 0.95 as keypoints. To track these keypoints, we employ a spatio-temporal nearest-neighbor strategy, which is designed to eliminate the drifts during the tracking process.

For each identified keypoint, we search for a neighboring keypoint within a spatial radius of 4 pixels and a temporal window of 12 milliseconds. If a neighbor is found, we associate the new keypoint with its existing track. In cases where multiple neighboring keypoints are found, we prioritize the closest one for correspondence. Conversely, if no neighboring keypoint is found, we initiate a new tracking sequence specifically from the new keypoint.

\subsubsection{Training set}
Following the approach in \cite{chiberre2022long}, we apply a series of homography transformations \{$\mathcal{T}^{(0,1)}, \mathcal{T}^{(0,2)}, ...,\mathcal{T}^{(0,{M_1)}}$\} to an initial image $I^0$ from COCO \cite{lin2014microsoft}. This yields an image sequence \{$I^1, I^2, ..., I^{M_1}$\}  that encompasses motion:  
\begin{equation}
    I^n = \mathrm{Warp}(I^0, \mathcal{T}^{(0,n)}), n \in \mathbb{Z},\, 1 \leq n \leq M_1.
\end{equation}

\begin{table*}[!htbp]
    \centering
    \setlength{\tabcolsep}{5.5pt}
    \caption{Performance comparison on our collected dataset. ``-" indicates cases with frame intervals exceeding $\delta t$.}
    \label{tab:compare}
    \begin{tabular}[t]{c|l|r|r|r|r|r|c}
        \toprule
        \toprule
         \multirow{2}{*}{Scene} & \multicolumn{1}{c|}{\multirow{2}{*}{Method}} & \multicolumn{1}{c|}{$\delta t=25 ms$} & \multicolumn{1}{c|}{$\delta t=50 ms$} & \multicolumn{1}{c|}{$\delta t=100 ms$} & \multicolumn{1}{c|}{$\delta t=150 ms$} & \multicolumn{1}{c|}{$\delta t=200 ms$} & \multirow{2}{*}{Track Time (s)$\uparrow$} \\
         & & RPE\,(RFM)$\downarrow$ & RPE\,(RFM)$\downarrow$ & RPE\,(RFM)$\downarrow$ & RPE\,(RFM)$\downarrow$ & RPE\,(RFM)$\downarrow$ \\
        \midrule
        \multirow{5}{*}{Overexposure}
          & Shi-Tomasi\cite{shi1994good}+LK\cite{lucas1981iterative} & 2.47\,\,(0.00) & 3.97\,\,(0.0) & 17.21\,\,(0.00) & 10.02\,\,(0.01) & 24.02\,\,(0.02) & \underline{3.38} \\
          & Superpoint\cite{detone2018superpoint} & 22.46\,\,(0.00) & 21.76\,\,(0.00) & 18.38\,\,(0.00) & 36.65\,\,(0.01) & 23.61\,\,(0.01) & \textbf{9.25}\\
          & \textcolor{gray}{Superpoint*\cite{detone2018superpoint}} & \textcolor{gray}{1.80\,\,(0.00)} & \textcolor{gray}{2.18\,\,(0.00)} & \textcolor{gray}{2.64\,\,(0.00)} & \textcolor{gray}{3.02\,\,(0.01)} & \textcolor{gray}{3.24\,\,(0.01)} & \textcolor{gray}{8.78}\\
          & Long-lived\cite{chiberre2022long} &  \underline{2.36}\,\,(0.05) & \underline{2.16}\,\,(0.19) & \underline{2.07}\,\,(0.40) & \underline{2.51}\,\,(0.52) & \textbf{1.82}\,\,(0.65) & 0.43\\
          & \textbf{FE-DeTr (Ours)} & \textbf{1.76}\,\,(0.00) & \textbf{1.90}\,\,(0.00) & \textbf{{2.01}}\,\,(0.01) & \textbf{2.10}\,\,(0.02) & \underline{{2.08}}\,\,(0.03) & 2.15\\
        \midrule
        \multirow{5}{*}{Dark}
          & Shi-Tomasi\cite{shi1994good}+LK\cite{lucas1981iterative} & 5.63\,\,(0.01) & 9.02\,\,(0.05) & 9.02\,\,(0.10) & 58.14\,\,(0.17) & 26.20\,\,(0.25) & \underline{1.63}\\
          & Superpoint\cite{detone2018superpoint} & 64.31\,\,(0.00) & 47.43\,\,(0.01) & 28.55\,\,(0.06) & 21.71\,\,(0.16) & 16.07\,\,(0.25) & \textbf{3.16}\\
          & \textcolor{gray}{Superpoint*\cite{detone2018superpoint}} & \textcolor{gray}{5.59\,\,(0.00)} & \textcolor{gray}{8.09\,\,(0.00)} & \textcolor{gray}{7.95\,\,(0.03)} & \textcolor{gray}{7.70\,\,(0.06)} & \textcolor{gray}{8.06\,\,(0.09)} & \textcolor{gray}{3.83}\\
          & Long-lived\cite{chiberre2022long} & \underline{2.28}\,\,(0.13) & \underline{2.06}\,\,(0.27) & \textbf{1.84}\,\,(0.53) & \textbf{1.71}\,\,(0.71) & \underline{2.11}\,\,(0.79) & 0.36\\
          & \textbf{FE-DeTr (Ours)} & \textbf{1.57}\,\,(0.03) & \textbf{1.86}\,\,(0.07) & \underline{1.89}\,\,(0.16) & \underline{2.12}\,\,(0.28) & \textbf{1.92}\,\,(0.39) & 0.91\\
        \midrule
        \multirow{5}{*}{HDR}
          & Shi-Tomasi\cite{shi1994good}+LK\cite{lucas1981iterative} & 1.75\,\,(0.01) & 2.64\,\,(0.02) & 4.17\,\,(0.04) & 5.72\,\,(0.06) & 8.20\,\,(0.08) & \underline{3.49}\\
          & Superpoint\cite{detone2018superpoint} & 8.01\,\,(0.00) & 8.61\,\,(0.00) & 16.52\,\,(0.01) & 10.66\,\,(0.03) & 9.19\,\,(0.05) & \textbf{9.01}\\
          & \textcolor{gray}{Superpoint*\cite{detone2018superpoint}} & \textcolor{gray}{1.67\,\,(0.00)} & \textcolor{gray}{1.97\,\,(0.00)} & \textcolor{gray}{2.32\,\,(0.00)} & \textcolor{gray}{2.51\,\,(0.01)} & \textcolor{gray}{2.83\,\,(0.01)} & \textcolor{gray}{8.71}\\
          & Long-lived\cite{chiberre2022long} & \underline{2.20}\,\,(0.08) & \underline{2.00}\,\,(0.16) & \underline{2.20}\,\,(0.35) & \underline{2.11}\,\,(0.53) & \textbf{1.74}\,\,(0.64) & 0.50\\
          & \textbf{FE-DeTr (Ours)} & \textbf{1.73}\,\,(0.00) & \textbf{1.86}\,\,(0.02) & \textbf{1.85}\,\,(0.06) & \textbf{1.81}\,\,(0.10) & \underline{1.75}\,\,(0.13) & 2.49\\
        \midrule
        \midrule
        \multirow{5}{*}{Blur}
          & Shi-Tomasi\cite{shi1994good}+LK\cite{lucas1981iterative} & \multicolumn{1}{c|}{-} & \multicolumn{1}{c|}{-} & 2.67\,\,(0.06) & \multicolumn{1}{c|}{-} & \underline{2.85}\,\,(0.10) & 1.49 \\
          & Superpoint\cite{detone2018superpoint} & \multicolumn{1}{c|}{-} & \multicolumn{1}{c|}{-} & 20.09\,\,(0.01) & \multicolumn{1}{c|}{-} & 20.05\,\,(0.08) & \textbf{3.93} \\
          & \textcolor{gray}{Superpoint*\cite{detone2018superpoint}} & \multicolumn{1}{c|}{\textcolor{gray}{-}} & \multicolumn{1}{c|}{\textcolor{gray}{-}} & \textcolor{gray}{5.31\,\,(0.01)} & \multicolumn{1}{c|}{\textcolor{gray}{-}} & \textcolor{gray}{11.41\,\,(0.09)} & \textcolor{gray}{2.24}\\
          & Long-lived\cite{chiberre2022long} & \underline{1.78}\,\,(0.01) & \textbf{1.74}\,\,(0.03) & \textbf{1.73}\,\,(0.05) & \textbf{1.71}\,\,(0.08) & \textbf{1.79}\,\,(0.16) & 0.74\\
          & \textbf{FE-DeTr (Ours)} & \textbf{1.60}\,\,(0.00) & \underline{1.96}\,\,(0.01) & \underline{2.67}\,\,(0.06) & \underline{2.85}\,\,(0.10) & 3.46\,\,(0.17) & \underline{1.49}\\
        \bottomrule
        \bottomrule
    \end{tabular}\vspace{-5mm}
\end{table*}

Subsequently, we simulate an event stream from this image sequence using an event simulator \cite{chiberre2022long}. 
We define $M_2 < M_1$ as the frame interval and uniformly divide the image sequence\{$I^1, I^2, ..., I^{M_1}$\} into intervals $\{ S^1=\{I^1, I^2, ..., I^{M_2}\}, S^2, ..., S^{{M_1} / {M_2}} \}$.
For each frame interval $S^n$, we select the last $M_3$ frames and calculate their average to obtain a blurred image $I$. 
Here, $M_3$ signifies the exposure time within the range $[0.2M_2, 0.6M_2]$.
Additionally, we apply random data augmentation to improve the robustness of model training, including random event noises and random grayscale transformations on blurred images.

\subsubsection{Training details}
\label{sec: Training details}
We configure our system with the following parameters: $B=10$ for event representation, $N=10$ for network outputs, $M=30$ for patch size, and employ cosine similarity as the distance metric for our loss function. We use truncated-backpropagation-through-time of 10-time steps and employ the Adam optimizer for the training process, which consists of two stages: 

\textbf{(a)} We first train the network for 30 epochs with an initial learning rate $lr$ of 0.0003 and an initial $\alpha$ of 0.25. 
At the 6th, 12th, and 18th epochs, we dynamically adjust $lr$ by reducing it to $75\%$ of its previous value while simultaneously increasing $\alpha$ by a factor of 2.
This stage is designed to enable the network to learn temporal motion relationships, thereby enhancing the consistency of its output detection. 

\textbf{(b)} We then train for 1 epoch with $lr = 0.0003$ and $\alpha=2.0$, utilizing the standard mask $\mathcal{M}$ representation in the Eq. (\ref{loss mask}). The primary objective is to suppress responses from homogeneous regions to reduce false positives.

\section{Experiments}
\label{sec:experiments}
In this section, we first introduce the collected dataset and the evaluation metrics, and then we compare the proposed method with the state-of-the-art image-based and event-based methods. Finally, we give the ablation study of FE-DeTr. 
\begin{figure}[t]
    \centering
    \includegraphics[width=1.0\linewidth]{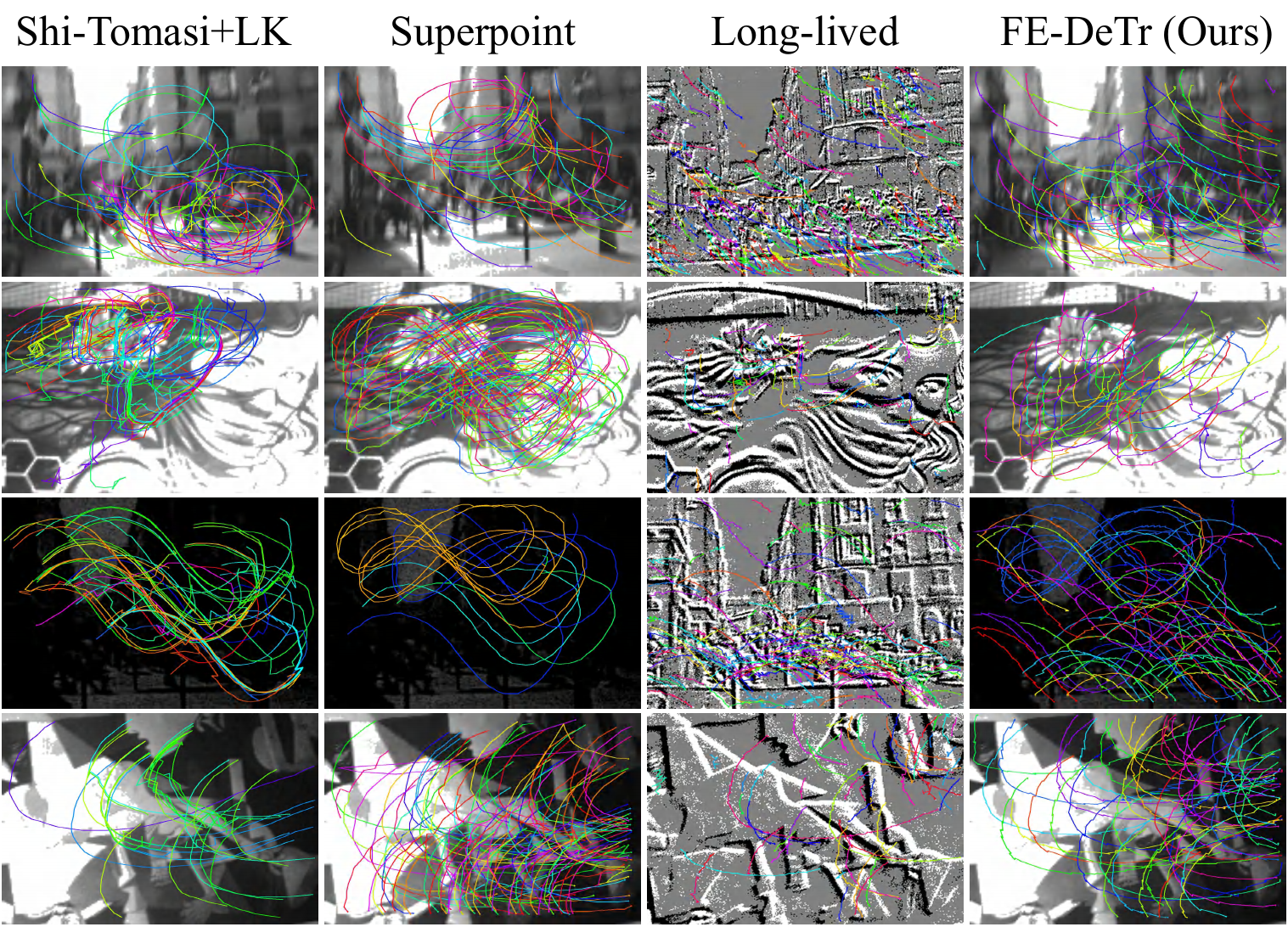}
    \caption{Tracking trajectories comparison under different conditions: Blur (1st row), overexposure (2nd row), dark (3rd row), and HDR (4th row).}
    \vspace{-5mm}
    \label{fig:compare}
\end{figure}

\subsection{Datasets and Metrics}
\label{new_dataset}
Since no publicly available keypoint dataset captures both image frames and event data under extreme conditions,
we create a new keypoint detection and tracking dataset, named Extreme Corner.
Following the setup of the HVGA ATIS Corner Dataset \cite{manderscheid2019speed}, we employ a DAVIS346 event camera to capture planar patterns of 4 distinct images to simplify the evaluation process. 
We set up three extreme lighting scenarios: \textbf{Overexposure}, \textbf{Dark}, and High Dynamic Range (\textbf{HDR}).
For each scenario, we collect an event stream of approximately 10 seconds in duration, accompanied by images captured at a frame interval of 25 milliseconds. To simulate \textbf{Blur} condition, we extend the frame interval and exposure time, and then record event streams and corresponding images at a 100 milliseconds frame interval, also for roughly 10 seconds. 
Under each extreme condition, we record two sequences for each planar scene,
culminating in a comprehensive dataset comprising 32 sequences. 

\begin{table*}[htbp]
    \centering
    \setlength{\tabcolsep}{5.5pt}
    \caption{Ablation study on the setup of the proposed FE-DeTr. ``-" indicates cases with frame intervals exceeding $\delta t$. ``NaN" means failed cases. ``/" indicates that the total number of tracking trajectories is less than 100.}
    \label{tab:abalation}
    \begin{tabular}[t]{c|l|r|r|r|r|r|c}
        \toprule
        \toprule
         \multirow{2}{*}{Scene} & \multicolumn{1}{c|}{\multirow{2}{*}{Setup}} & \multicolumn{1}{c|}{$\delta t=25 ms$} & \multicolumn{1}{c|}{$\delta t=50 ms$} & \multicolumn{1}{c|}{$\delta t=100 ms$} & \multicolumn{1}{c|}{$\delta t=150 ms$} & \multicolumn{1}{c|}{$\delta t=200 ms$} & \multirow{2}{*}{Track Time (s)$\uparrow$} \\
         & & RPE\,(RFM)$\downarrow$ & RPE\,(RFM)$\downarrow$ & RPE\,(RFM)$\downarrow$ & RPE\,(RFM)$\downarrow$ & RPE\,(RFM)$\downarrow$ \\
       \midrule
       \midrule
       \multirow{5}{*}{Extreme Lighting}
          & (a). w/o $L_{cp}$ & 2.17\,\,(0.00) & 2.55\,\,(0.00) & 2.93\,\,(0.01) & 3.07\,\,(0.03) & 3.66\,\,(0.04) & \textbf{2.27}\\
          & (b). w/o MAH & 2.58\,\,(0.01) & 4.19\,\,(0.04) & 3.13\,\,(0.13) & 4.98\,\,(0.22) & 3.13\,\,(0.31) & 0.93\\
          & (c). w/o event & 2.47\,\,(0.42) & 8.08\,\,(0.83) & \underline{2.03}\,\,(0.93) & NaN\,\,(1.00) & NaN\,\,(1.00) & / \\
          & (d). w/o frame & \textbf{1.30}\,\,(0.39) & \textbf{1.59}\,\,(0.55) & 2.26\,\,(0.75) & \underline{2.24}\,\,(0.88) & 5.12\,\,(0.94) & 0.29\\
          & (e). full & \underline{1.69}\,\,(0.01) & \underline{1.87}\,\,(0.03) & \textbf{1.92}\,\,(0.08) & \textbf{2.00}\,\,(0.13) & \textbf{1.92}\,\,(0.18) & \underline{1.85}\\
        \midrule
       \midrule
        \multirow{5}{*}{Blur}
          & (a). w/o $L_{cp}$ & 1.73\,\,(0.00) & 2.23\,\,(0.00) & 3.34\,\,(0.00) & 3.92\,\,(0.02) & 4.22\,\,(0.06) & \textbf{1.93}\\
          & (b). w/o MAH & 2.37\,\,(0.00) & 3.01\,\,(0.01) & 4.29\,\,(0.09) & 5.96\,\,(0.21) & 4.76\,\,(0.30) & 0.91\\
          & (c). w/o event & \multicolumn{1}{c|}{-} & \multicolumn{1}{c|}{-} & 5.07\,\,(0.40) & \multicolumn{1}{c|}{-} & NaN\,\,(1.00) & 0.41\\
          & (d). w/o frame & \textbf{1.34}\,\,(0.08) & \textbf{1.55}\,\,(0.13) & \textbf{1.81}\,\,(0.27) & \textbf{2.13}\,\,(0.45) & \textbf{2.60}\,\,(0.60) & 0.52\\
          & (e). full & \underline{1.60}\,\,(0.00) & \underline{1.96}\,\,(0.01) & \underline{2.67}\,\,(0.06) & \underline{2.85}\,\,(0.10) & \underline{3.46}\,\,(0.17) & \underline{1.49}\\
        \bottomrule
        \bottomrule
    \end{tabular}\vspace{-5mm}
\end{table*}

We use two key metrics for evaluation: the $\delta t$-homography Reprojection Error (RPE) for assessing detection accuracy, and the average Track Time for examining detection stability.
Definitions of these metrics can be found in \cite{chiberre2022long}. Alongside the RPE, we report the Ratio of Failed Matches (RFM), which indicates the ratio of failed matches to total matches, thereby reflecting the reliability of RPE.

\subsection{Result Comparisons}
To demonstrate the effectiveness of our method, we select several traditional and SOTA learning-based methods using either image frames or event streams for comparison. (1) Shi-Tomasi \cite{shi1994good} + Lucas-Kanade \cite{lucas1981iterative} represents a traditional frame-based keypoint detection and optical flow method. (2) Superpoint \cite{detone2018superpoint} represents a learning-based keypoint detection and description method also on frames. (3) Long-lived \cite{chiberre2022long} is a state-of-the-art event-based keypoint detection and tracking method. To ensure a fair comparison, all methods refrain from utilizing enhancements such as image pyramids or feature pyramids. 
Parameters for all methods are adapted to achieve the best performance on the same dataset.

Superpoint generates numerous mismatches on regular chessboard sequences because corners on the chessboard exhibit extremely high similarity. Therefore, we also report its performance on the proposed dataset without the chessboard sequences, denoted as \textcolor{gray}{Superpoint*} in Table \ref{tab:compare} for reference. 

As shown in Table \ref{tab:compare}, the proposed FE-DeTr achieves the best accuracy under extreme lighting conditions, while also maintaining robust performance in terms of accuracy, failure rate, and tracking time under blur conditions. 
Frame-based methods face challenges due to the absence of mechanisms to filter out mismatched points, leading to protracted tracking times and a concomitant decrease in accuracy.
The event-based method Long-lived \cite{chiberre2022long} achieves good accuracy in low-noise and high-speed motion scenes (Blur). However, its stability is compromised under extreme lighting conditions, reflected by its high RFM and extremely low Track Time.
This performance degradation is especially pronounced in dark (low light) scenes filled with event noise. 
By integrating the complementary information of image frames and event data, the proposed FE-DeTr achieves a balance between accuracy and stability.


Furthermore, we qualitatively plot the tracking trajectories of different methods in Fig. \ref{fig:compare}. On the one hand, the proposed FE-DeTr exhibits smoother trajectories compared to frame-based methods. On the other hand, our method achieves longer tracking trajectories compared to event-based methods. These results further demonstrate the effectiveness of our method under various challenging conditions.

\subsection{Ablation Study}

\begin{figure}[t]
    \centering
    \includegraphics[width=1.0\linewidth]{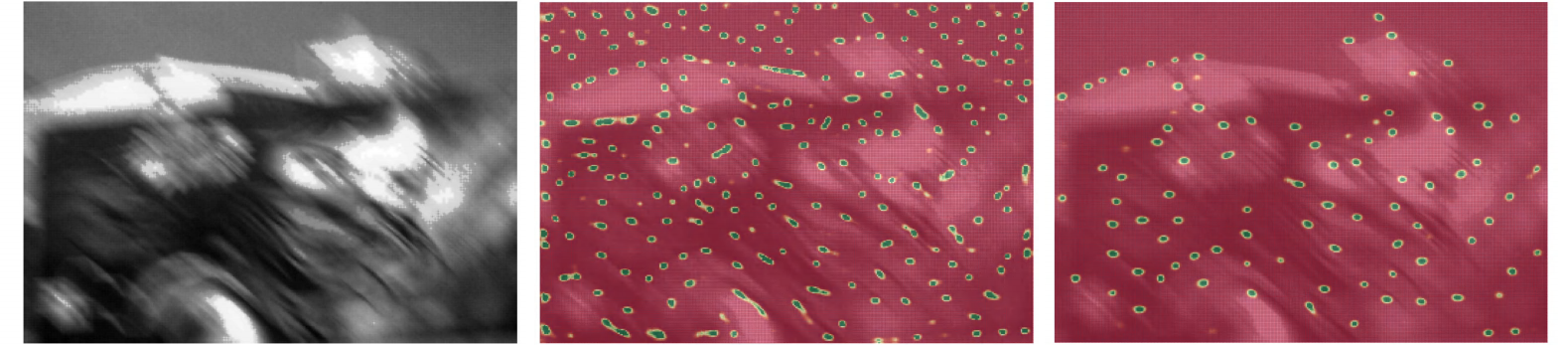}
    \caption{Reference image frame (left); output heatmap after training without $L_{cp}$ (middle); output heatmap after $L_{cp}$ supervision (right).}
    \vspace{-5mm}
    \label{fig:withou_Lcp}
\end{figure}

$\textbf{Impact of Consistency Peaky Loss}$.
To validate the effectiveness of the proposed Consistency Peaky Loss in eliminating false positives, we present the results of FE-DeTr without $L_{cp}$ in row (a) of Table \ref{tab:abalation}. 
Compared with the full model in row (e),
it can be observed that the absence of this loss increases false responses, leading to a greater number of incorrect point matches. While this extends tracking lifetimes, it also decreases accuracy.
Fig. 
\ref{fig:withou_Lcp} shows a comparison between the output heatmaps generated by a model trained without $L_{cp}$ loss versus the full model.

\textbf{Impact of MAH}.
We verify the effectiveness of the MAH module by replacing the deformable convolution with regular convolution and removing the iterative steps. The results are shown in row (b) of Table \ref{tab:abalation}.
Compared with the full model in row (e), we can find a significant performance drop after the modification, highlighting the strong capability of MAH in utilizing motion information to improve the response’s temporal consistency. 

\textbf{Impact of Input Modalities}.
To validate the effectiveness of fusing images and events, we test the FE-DeTr on a single modality with the other input to 0. The results of rows (c) and (d) of each scene in Table \ref{tab:abalation} show the performance on images and events, respectively. 
When using only the frame input, a noticeable performance drop can be observed. 
This is because using only the frame modality can't generate the high-temporal-resolution responses required for tracking.
When utilizing only the event modality, tracking accuracy is improved, indicating that the keypoint localization accuracy is somewhat influenced by low-quality images.
However, the higher RFM and extremely short tracking time suggest poor stability when only using events, especially in extreme lighting conditions with a high noise level. The fusion of image frames significantly improves the stability of detection and tracking.


\section{Conclusions and Future Work}
\label{sec:conclusion}
In this paper, we propose FE-DeTr, a keypoint detection and tracking method that integrates both images and events. The keypoint detection network is supervised based on the relative motion relationships at different time instants, enabling the exploitation of complementary information between two modalities. Compared to prior arts, FE-DeTr achieves the best comprehensive performance with high localization accuracy and stable tracking duration. Future work will focus on applying FE-DeTr to the downstream SLAM task.


\section{Acknowledgement}
This work was supported by NSFC grants under Grant 62301370 and 62271354, the Natural Science Foundation of Hubei Province, China under Grant 2022CFB600.

\bibliographystyle{IEEEtran.bst}
\bibliography{references}

\begin{thebibliography}{10}
\providecommand{\url}[1]{#1}
\csname url@rmstyle\endcsname
\providecommand{\newblock}{\relax}
\providecommand{\bibinfo}[2]{#2}
\providecommand\BIBentrySTDinterwordspacing{\spaceskip=0pt\relax}
\providecommand\BIBentryALTinterwordstretchfactor{4}
\providecommand\BIBentryALTinterwordspacing{\spaceskip=\fontdimen2\font plus
\BIBentryALTinterwordstretchfactor\fontdimen3\font minus
  \fontdimen4\font\relax}
\providecommand\BIBforeignlanguage[2]{{%
\expandafter\ifx\csname l@#1\endcsname\relax
\typeout{** WARNING: IEEEtran.bst: No hyphenation pattern has been}%
\typeout{** loaded for the language `#1'. Using the pattern for}%
\typeout{** the default language instead.}%
\else
\language=\csname l@#1\endcsname
\fi
#2}}

\bibitem{detone2018superpoint}
D.~DeTone, T.~Malisiewicz, and A.~Rabinovich, ``Superpoint: Self-supervised
  interest point detection and description,'' in \emph{Proceedings of the IEEE
  Conference on Computer Vision and Pattern Recognition Workshops}, 2018, pp.
  224--236.

\bibitem{christiansen2019unsuperpoint}
P.~H. Christiansen, M.~F. Kragh, Y.~Brodskiy, and H.~Karstoft, ``Unsuperpoint:
  End-to-end unsupervised interest point detector and descriptor,'' \emph{arXiv
  preprint arXiv:1907.04011}, 2019.

\bibitem{revaud2019r2d2}
J.~Revaud, P.~Weinzaepfel, C.~D. Souza, and M.~Humenberger, ``R2d2: Repeatable
  and reliable detector and descriptor,'' in \emph{Proceedings of the
  International Conference on Neural Information Processing Systems}, 2019, p.
  12414–12424.

\bibitem{dosovitskiy2015flownet}
A.~Dosovitskiy, P.~Fischer, E.~Ilg, P.~Hausser, C.~Hazirbas, V.~Golkov, P.~Van
  Der~Smagt, D.~Cremers, and T.~Brox, ``Flownet: Learning optical flow with
  convolutional networks,'' in \emph{Proceedings of the IEEE International
  Conference on Computer Vision}, 2015, pp. 2758--2766.

\bibitem{teed2020raft}
Z.~Teed and J.~Deng, ``Raft: Recurrent all-pairs field transforms for optical
  flow,'' in \emph{European Conference on Computer Vision}, 2020, pp. 402--419.

\bibitem{tian2017l2}
Y.~Tian, B.~Fan, and F.~Wu, ``L2-net: Deep learning of discriminative patch
  descriptor in euclidean space,'' in \emph{Proceedings of the IEEE Conference
  on Computer Vision and Pattern Recognition}, 2017, pp. 661--669.

\bibitem{jiang2021cotr}
W.~Jiang, E.~Trulls, J.~Hosang, A.~Tagliasacchi, and K.~M. Yi, ``Cotr:
  Correspondence transformer for matching across images,'' in \emph{Proceedings
  of the IEEE International Conference on Computer Vision}, 2021, pp.
  6207--6217.

\bibitem{doersch2022tap}
C.~Doersch, A.~Gupta, L.~Markeeva, A.~Recasens, L.~Smaira, Y.~Aytar,
  J.~Carreira, A.~Zisserman, and Y.~Yang, ``Tap-vid: A benchmark for tracking
  any point in a video,'' \emph{Advances in Neural Information Processing
  Systems}, vol.~35, pp. 13\,610--13\,626, 2022.

\bibitem{harley2022particle}
A.~W. Harley, Z.~Fang, and K.~Fragkiadaki, ``Particle video revisited: Tracking
  through occlusions using point trajectories,'' in \emph{European Conference
  on Computer Vision}, 2022, pp. 59--75.

\bibitem{doersch2023tapir}
C.~Doersch, Y.~Yang, M.~Vecerik, D.~Gokay, A.~Gupta, Y.~Aytar, J.~Carreira, and
  A.~Zisserman, ``Tapir: Tracking any point with per-frame initialization and
  temporal refinement,'' \emph{arXiv preprint arXiv:2306.08637}, 2023.

\bibitem{zhao2022balf}
Z.~Zhao, Y.~Zhai, B.~M. Chen, and P.~Liu, ``Balf: Simple and efficient blur
  aware local feature detector,'' \emph{arXiv preprint arXiv:2211.14731}, 2022.

\bibitem{gallego2020event}
G.~Gallego, T.~Delbr{\"u}ck, G.~Orchard, C.~Bartolozzi, B.~Taba, A.~Censi,
  S.~Leutenegger, A.~J. Davison, J.~Conradt, K.~Daniilidis, \emph{et~al.},
  ``Event-based vision: A survey,'' \emph{IEEE Transactions on Pattern Analysis
  and Machine Intelligence}, vol.~44, no.~1, pp. 154--180, 2020.

\bibitem{zhang2023generalizing}
X.~Zhang, L.~Yu, W.~Yang, J.~Liu, and G.-S. Xia, ``Generalizing event-based
  motion deblurring in real-world scenarios,'' in \emph{Proceedings of the
  IEEE/CVF International Conference on Computer Vision}, 2023, pp.
  10\,734--10\,744.

\bibitem{chen2024motion}
K.~Chen and L.~Yu, ``Motion deblur by learning residual from events,''
  \emph{IEEE Transactions on Multimedia}, 2024.

\bibitem{vasco2016fast}
V.~Vasco, A.~Glover, and C.~Bartolozzi, ``Fast event-based harris corner
  detection exploiting the advantages of event-driven cameras,'' in \emph{IEEE
  International Conference on Intelligent Robots and Systems}, 2016, pp.
  4144--4149.

\bibitem{mueggler2017fast}
E.~Mueggler, C.~Bartolozzi, and D.~Scaramuzza, ``Fast event-based corner
  detection,'' in \emph{British Machine Vision Conference}, 2017.

\bibitem{alzugaray2018asynchronous}
I.~Alzugaray and M.~Chli, ``Asynchronous corner detection and tracking for
  event cameras in real time,'' \emph{IEEE Robotics and Automation Letters},
  vol.~3, no.~4, pp. 3177--3184, 2018.

\bibitem{li2019fa}
R.~Li, D.~Shi, Y.~Zhang, K.~Li, and R.~Li, ``Fa-harris: A fast and asynchronous
  corner detector for event cameras,'' in \emph{IEEE International Conference
  on Intelligent Robots and Systems}, 2019, pp. 6223--6229.

\bibitem{manderscheid2019speed}
J.~Manderscheid, A.~Sironi, N.~Bourdis, D.~Migliore, and V.~Lepetit, ``Speed
  invariant time surface for learning to detect corner points with event-based
  cameras,'' in \emph{Proceedings of the IEEE Conference on Computer Vision and
  Pattern Recognition}, 2019, pp. 10\,245--10\,254.

\bibitem{chiberre2021detecting}
P.~Chiberre, E.~Perot, A.~Sironi, and V.~Lepetit, ``Detecting stable keypoints
  from events through image gradient prediction,'' in \emph{Proceedings of the
  IEEE Conference on Computer Vision and Pattern Recognition}, 2021, pp.
  1387--1394.

\bibitem{chiberre2022long}
C.~Philippe, P.~Etienne, S.~Amos, and L.~Vincent, ``Long-lived accurate
  keypoints in event streams,'' \emph{arXiv preprint arXiv: 2209.10385}, 2022.

\bibitem{zhu2017event}
A.~Z. Zhu, N.~Atanasov, and K.~Daniilidis, ``Event-based feature tracking with
  probabilistic data association,'' in \emph{IEEE International Conference on
  Robotics and Automation}, 2017, pp. 4465--4470.

\bibitem{gehrig2018asynchronous}
D.~Gehrig, H.~Rebecq, G.~Gallego, and D.~Scaramuzza, ``Asynchronous,
  photometric feature tracking using events and frames,'' in \emph{European
  Conference on Computer Vision}, 2018, pp. 750--765.

\bibitem{messikommer2023data}
N.~Messikommer, C.~Fang, M.~Gehrig, and D.~Scaramuzza, ``Data-driven feature
  tracking for event cameras,'' in \emph{Proceedings of the IEEE Conference on
  Computer Vision and Pattern Recognition}, 2023, pp. 5642--5651.

\bibitem{yu2023detecting}
H.~Yu, H.~Li, W.~Yang, L.~Yu, and G.-S. Xia, ``Detecting line segments in
  motion-blurred images with events,'' \emph{IEEE Transactions on Pattern
  Analysis and Machine Intelligence}, 2023.

\bibitem{yi2016lift}
K.~M. Yi, E.~Trulls, V.~Lepetit, and P.~Fua, ``Lift: Learned invariant feature
  transform,'' in \emph{European Conference on Computer Vision}, 2016, pp.
  467--483.

\bibitem{gleize2023silk}
P.~Gleize, W.~Wang, and M.~Feiszli, ``Silk--simple learned keypoints,''
  \emph{arXiv preprint arXiv:2304.06194}, 2023.

\bibitem{lucas1981iterative}
B.~D. Lucas and T.~Kanade, ``An iterative image registration technique with an
  application to stereo vision,'' in \emph{International Joint Conference on
  Artificial Intelligence}, vol.~2, 1981, pp. 674--679.

\bibitem{sun2018pwc}
D.~Sun, X.~Yang, M.-Y. Liu, and J.~Kautz, ``Pwc-net: Cnns for optical flow
  using pyramid, warping, and cost volume,'' in \emph{Proceedings of the IEEE
  Conference on Computer Vision and Pattern Recognition}, 2018, pp. 8934--8943.

\bibitem{zhu2019unsupervised}
A.~Z. Zhu, L.~Yuan, K.~Chaney, and K.~Daniilidis, ``Unsupervised event-based
  learning of optical flow, depth, and egomotion,'' in \emph{Proceedings of the
  IEEE Conference on Computer Vision and Pattern Recognition}, 2019, pp.
  989--997.

\bibitem{guo2022cmt}
J.~Guo, K.~Han, H.~Wu, Y.~Tang, X.~Chen, Y.~Wang, and C.~Xu, ``Cmt:
  Convolutional neural networks meet vision transformers,'' in
  \emph{Proceedings of the IEEE Conference on Computer Vision and Pattern
  Recognition}, 2022, pp. 12\,175--12\,185.

\bibitem{shi2015convolutional}
X.~Shi, Z.~Chen, H.~Wang, D.-Y. Yeung, W.-K. Wong, and W.-c. Woo,
  ``Convolutional lstm network: A machine learning approach for precipitation
  nowcasting,'' in \emph{Proceedings of the International Conference on Neural
  Information Processing Systems}, 2015, pp. 802--810.

\bibitem{zhang2023frame}
J.~Zhang, Y.~Wang, W.~Liu, M.~Li, J.~Bai, B.~Yin, and X.~Yang, ``Frame-event
  alignment and fusion network for high frame rate tracking,'' in
  \emph{Proceedings of the IEEE Conference on Computer Vision and Pattern
  Recognition}, 2023, pp. 9781--9790.

\bibitem{hu2018squeeze}
J.~Hu, L.~Shen, and G.~Sun, ``Squeeze-and-excitation networks,'' in
  \emph{Proceedings of the IEEE Conference on Computer Vision and Pattern
  Recognition}, 2018, pp. 7132--7141.

\bibitem{zhang2018learning}
L.~Zhang and S.~Rusinkiewicz, ``Learning to detect features in texture
  images,'' in \emph{Proceedings of the IEEE Conference on Computer Vision and
  Pattern Recognition}, 2018, pp. 6325--6333.

\bibitem{lin2014microsoft}
T.-Y. Lin, M.~Maire, S.~Belongie, J.~Hays, P.~Perona, D.~Ramanan,
  P.~Doll{\'a}r, and C.~L. Zitnick, ``Microsoft coco: Common objects in
  context,'' in \emph{European Conference on Computer Vision}, 2014, pp.
  740--755.

\bibitem{shi1994good}
J.~Shi and C.~Tomasi, ``Good features to track,'' in \emph{Proceedings of the
  IEEE Conference on Computer Vision and Pattern Recognition}, 1994, pp.
  593--600.

\end{thebibliography}

\end{document}